\documentclass[letterpaper, 10 pt, conference]{ieeeconf}  

\IEEEoverridecommandlockouts                              

\usepackage{graphics} 
\usepackage{epsfig} 
\usepackage{mathptmx} 
\usepackage{times} 
\usepackage{amsmath} 
\usepackage{amssymb}  
\usepackage{booktabs}  
\usepackage{multirow}  
\graphicspath{ {./images/} }
\title{\LARGE \bf
SGN-CIRL: Scene Graph-based Navigation with Curriculum, Imitation, and Reinforcement Learning
}
\author{Nikita Oskolkov$^{1}$, Huzhenyu Zhang$^{1}$, Dmitry Makarov$^{1,2}$, Dmitry Yudin$^{1,3}$, Aleksandr Panov$^{1,3}$
\thanks{$^{1}$Nikita Oskolkov, Huzhenyu Zhang, Dmitry Makarov, Dmitry Yudin, and  Aleksandr Panov are with Moscow Institute of Physics and Technology (MIPT), Dolgoprudny, Russia
{\tt\small oskolkov.ns@phystech.edu, jianghuzhenyu@gmail.com, makarov.da@mipt.ru, yudin.da@mipt.ru, panov.ai@mipt.ru}}%
\thanks{$^{2}$Dmitry Makarov is with Federal Research Center for Computer Science and Control of Russian Academy of Sciences, Moscow, Russia 
{\tt\small makarov@isa.ru}}%
\thanks{$^{3}$ Dmitry Yudin and Aleksandr Panov are also with Artificial Intelligence Research Institute (AIRI), Moscow, Russia
{\tt\small \{yudin.da,panov\}@airi.net}}%
}

\begin{document}
\maketitle




\begin{abstract}
The 3D scene graph models spatial relationships between objects, enabling the agent to efficiently navigate in a partially observable environment and predict the location of the target object.
This paper proposes an original framework named \textbf{SGN-CIRL} (\textbf{3D Scene Graph-Based Reinforcement Learning Navigation}) for mapless reinforcement learning-based robot navigation with learnable representation of open-vocabulary 3D scene graph. 
To accelerate and stabilize the training of reinforcement learning-based algorithms, the framework also employs imitation learning and curriculum learning. 
The first one enables the agent to learn from demonstrations, while the second one structures the training process by gradually increasing task complexity from simple to more advanced scenarios.  
Numerical experiments conducted in the Isaac Sim environment showed that using a 3D scene graph for reinforcement learning significantly increased the success rate in difficult navigation cases. 
The code is open-sourced and available at: https://github.com/Xisonik/Aloha\_graph.

\end{abstract}

\section{INTRODUCTION}
Autonomous navigation is one of the key challenges in mobile robotics. Modern approaches often rely on localization and mapping \cite{chen2002robot} and \cite{Li2023Autonomous}. However, mapless navigation methods represent a promising alternative, allowing robots to navigate in dynamic environments without explicit mapping \cite{lee2022mobile} and \cite{mackay2022rl}. In such scenarios, the agent must make decisions based solely on its sensor observations and knowledge of the spatial structure of the surrounding environment.


Reinforcement Learning (RL) is actively used for navigation tasks due to its ability to optimize motion strategies based on accumulated experience. In this work, the agent is trained to navigate in an environment with a continuous action space consisting of linear and angular velocity. However, classical RL methods may encounter challenges such as sparse rewards, unstable learning, and prolonged adaptation to complex conditions. To accelerate the training process and enhance its stability, this work employs Imitation Learning (IL) and Curriculum Learning (CL). IL enables the agent to leverage demonstration behavior in early and complex stages, reducing the likelihood of ineffective actions. CL introduces dynamically adjustable training complexity, starting with simple tasks (short distances, simple angles) and gradually increasing the difficulty.

\begin{figure}[!t]
    \centering
    \includegraphics[width=0.5\textwidth]{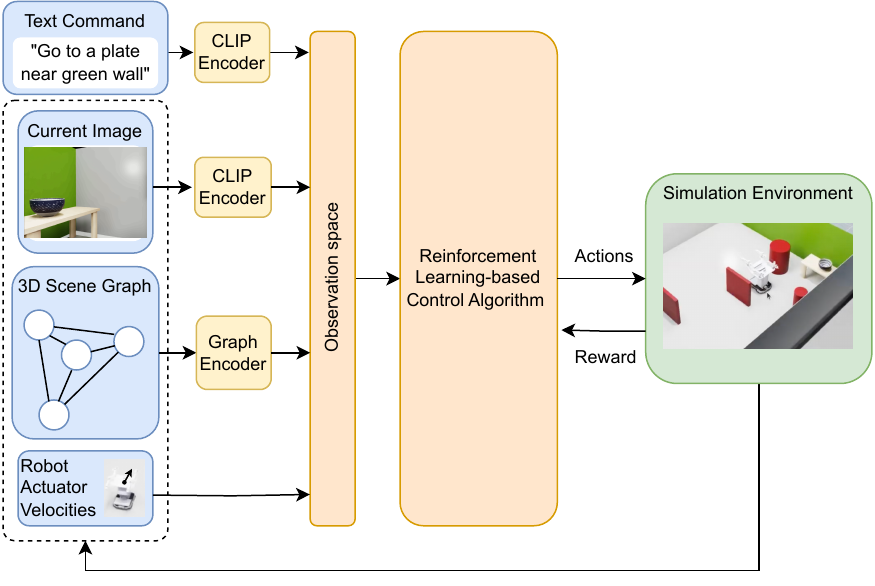}
    \caption{ Simplified diagram of the developed SGN-CIRL approach. We explore the possibility of using a 3D scene graph for robot navigation based on a reinforcement learning algorithm in a photorealistic simulation environment.
    }
    \label{fig:graphical_abstract}
\end{figure}

Additionally, the agent utilizes scene graphs to model spatial relationships between objects in the environment. This approach allows the robot to predict the most likely location of the target object and optimize its route, which is particularly useful in environments with complex structures. This paper presents a method named SGN-CIRL (see Table~\ref{fig:graphical_abstract}) for mapless navigation of a wheeled robot using RL, IL, CL, and scene graphs. 

The main contributions of this work are: 
\begin{itemize}
    \item We developed robot navigation approach which combines  reinforcement learning, imitation learning, and curriculum learning for accelerated and stable intelligent agent training;
    \item We proposed a learnable representation of open-vocabulary 3D scene graph that allowed a reinforcement learning-based navigation algorithm named SGN-CIRL to improve the success rate in a partially observable environment.
\end{itemize}


\section{Related Works}

\subsection{3D Scene Graph Construction}
3D scene graphs can provide the agent with rich scene knowledge, which could improve the agent's task performance. In general, there are two types of scene graph construction: one is real-time, and the other is pre-building. In \cite{singh2023scene}, the authors integrate the scene graph representation as an auxiliary signal for agents during the training of RL policies. The graph encoder is composed of three Graph Attention Layers \cite{velivckovic2018graph}. \cite{yin2025sg} constructs a 3D scene graph online and keeps updating and pruning it at each time step.

There are also works \cite{werby2024hierarchical}\cite{yan2024dynamic} that construct static 3D scene graphs before the agent performs tasks. These graphs often contain the location, size, and semantic information of objects. The prior knowledge provided by these 3D scene graphs could guide the agent to better perform given tasks, such as navigation or mobile manipulation.

ConceptGraph \cite{gu2024conceptgraphs} and BBQ \cite{linok2024barequeriesopenvocabularyobject} are advanced 3D scene graph construction algorithms. By abstracting objects and their relationships in the scene into a high-level graph structure, they effectively capture the semantic structure of the scene. The strength of these 3D scene graphs lies in their powerful object-level semantic reasoning capabilities, which integrate low-level geometric information with high-level semantic information, thereby enabling intelligent agents to achieve a richer understanding of the scene.

\subsection{RL Navigation with Indoor Scene Graph}
The study \cite{du2020learning} focuses on target-driven visual navigation, where an agent must navigate in an environment using only its visual observations. The authors propose three key methods to improve navigation quality: 
Object Relation Graph (ORG) – a graph that models probable spatial relationships between objects. This graph helps the agent constrain the search area for the target object, thereby improving navigation efficiency. Trial-driven Imitation Learning (IL) – reinforcement learning is complemented by imitation learning, which is used to create a more robust navigation policy. Tentative Policy Network (TPN) – a memory-augmented policy network that helps the agent escape dead-end states during testing. It uses external memory to detect looping (by comparing visual observations) and internal memory to store state-action pairs. The work shows that combining RL, knowledge graphs, and memory significantly enhances the agent's navigation in unknown environments.

A method for semantic navigation, based on the use of prior knowledge about spatial relationships between objects, is proposed in \cite{yang2018visual}. The authors integrated Graph Convolutional Networks (GCNs) into a deep reinforcement learning architecture (A3C), enabling the agent to account for object relationships and improve target search. Results obtained in the AI2-THOR simulator show that using knowledge graphs significantly improves navigation metrics, including success rates and route efficiency (SPL).

The paper \cite{xue2022using} presents a mapless navigation method for wheeled robots in warehouse environments, based on deep reinforcement learning and automatic curriculum learning. Unlike traditional planning methods, their approach uses the Soft Actor-Critic (SAC) algorithm to directly control the robot's velocities from sensor data (RGB camera and LiDAR), avoiding explicit map construction. To accelerate and stabilize training, the authors developed NavACL-Q—an enhanced version of Automatic Curriculum Learning (ACL)—which dynamically adjusts task difficulty, starting with simple episodes and gradually increasing their complexity. This approach reduces the likelihood of early agent stagnation and improves generalization. The work also incorporates elements of imitation learning (IL) through self-supervision: successful agent trajectories are used to adapt future tasks, similar to learning from a mentor.  Results in NVIDIA Isaac Sim demonstrates that our trained agent significantly outperforms the map-based navigation pipeline provided by NVIDIA Isaac Sim with an increased agent-goal distance and a wider initial relative agent-goal rotation.

The paper \cite{pfeiffer2018reinforced} deals with a Reinforced Imitation Learning method for mapless navigation of wheeled robots, combining imitation learning (IL) and reinforcement learning. In the first stage, the agent learns to mimic expert actions, and then optimizes its strategy using RL with Constrained Policy Optimization (CPO) to enhance safety. This approach significantly reduces training time compared to pure RL and improves model generalization compared to IL. The agent's inputs include LiDAR scans and the relative position of the target, while the output space consists of linear and angular velocities, making the method applicable to real-world mobile robotics scenarios. Experiments conducted in simulation and on a real TurtleBot2 robot show that accelerates training by approximately 5 times compared to RL without demonstrations and produces smoother and safer trajectories. However, the work does not utilize knowledge graphs.

A method for autonomous navigation of mobile robots in unknown environments, based on reinforcement learning and curriculum learning, is proposed in \cite{yin2024autonomous}. The Soft Actor-Critic algorithm is used as the RL method, and for adaptive task complexity, a Curriculum-based Energy Prioritization (CEP) mechanism is introduced, allowing the agent to first train on simple trajectories and then gradually increase difficulty. Additionally, a Fuzzy Logic Controller (FLC) is applied to enhance safety during movement. Experiments conducted in simulation and on a real TurtleBot3 robot show that using CL reduces training time and decreases the number of collisions compared to RL without curriculum learning. The work does not employ imitation learning or knowledge graphs, but the proposed approach demonstrates the effectiveness of combining CL and RL for mapless navigation.

Unlike the reviewed works, our approach combines IL, which helps avoid local minima, curriculum learning to gradually reduce dependence on expert demonstrations and increase the contribution of independent exploration, and a knowledge graph that provides spatial knowledge about the environment. This helps the agent better adapt to previously unseen environments and reduces the risk of overfitting to expert trajectories. Experiments show that this combination improves navigation success rates and efficiency metrics. Notably, in our problem setup, there are no additional sensors besides a camera, which significantly complicates the task. The novelty of our work lies in the proposed graph and its integration with IL (where expert data is obtained using Dijkstra's algorithm and heuristic control) and CL.

\begin{figure*}[!t]
    \centering
    \includegraphics[width=1\textwidth]{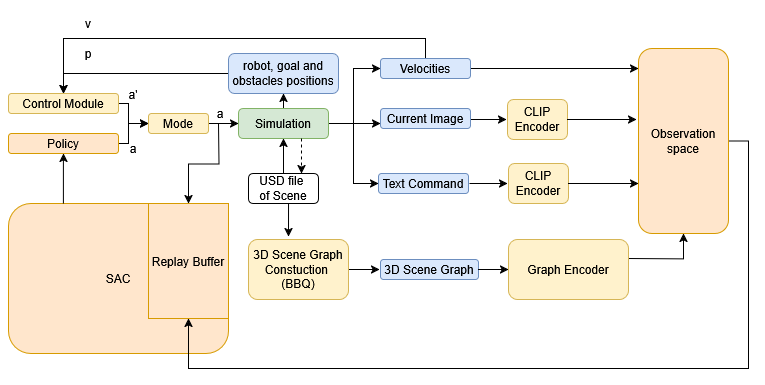}
    \caption{Scheme of the proposed SGN-CIRL approach for mapless image-based robot navigation in a simulation environment. Its main feature is the combination of reinforcement learning, curriculum learning, imitation learning implemented through the control module, and 3D scene graph construction. Here, v - current robot velocity, p - robot, goal and obstacles positions, a' - actions from the control module, a - actions from the policy.}
    \label{fig:block}
\end{figure*}

\section{SGN-CIRL Approach}
This section describes the module implementation details of the proposed SGN-CIRL approach shown in Fig.\ref{fig:block}.
\subsection{3D Scene Graph Construction and Encoding}
\textbf{3D Scene Graph}
We construct an indoor scene in Isaac Sim for the agent's navigation task. A moving camera is utilized to capture an RGB-D image sequence at a resolution of $1280 \times 720$ within the simulation environment. Various algorithms, such as ConceptGraph \cite{gu2024conceptgraphs} and BBQ \cite{linok2024barequeriesopenvocabularyobject}, are employed to construct the 3D scene graph. For each object in the scene, we extract a text description, along with the size and position of its 3D bounding box, which serve as part of the input for the RL policy. The 3D bounding box is represented by its extent (3 dimensions) and position (3 dimensions). Additionally, the text description is encoded using the CLIP model to generate a 512-dimensional embedding. These features for each object are concatenated to form a 518-dimensional tensor. The 3D scene graph is thus represented as a tensor of shape $(\text{num\_objects}, 518)$, which is first processed by a Graph Encoder before being fed into the reinforcement learning policy. The Graph Encoder is utilized to compress the direct output of the 3D scene graph algorithms. For each object in the scene, we compute the cosine similarity between the CLIP embedding of the object's textual description and the CLIP embedding of the target object-bowl pair. Subsequently, we compress the original 3D scene graph into a 512-dimensional vector through a pooling operation, with the CLIP cosine similarity values serving as the weights in this process.

\subsection{Explanation of the reinforcement learning approach}
\textbf{Problem statement}
The task of the navigation algorithm $f$ for a mobile robot is to reach the target position by selecting angular and linear velocities at each moment of movement while minimizing the cost of the robot's motion. The interaction process with the environment represents a Markov Decision Process (MDP), which requires defining \textbf{continuous}:

\begin{enumerate}
    \item \textbf{State space} $\mathcal{S}$:
    \begin{enumerate}
    \item Camera data  $X_{cam}$ at a resolution of $(400\times400\times3)$,
    \item Text command $X_{goal}$,
    \item Base velocity $V_{current}$ (2 DoF): the velocity consists of two components: linear and angular.
\end{enumerate}
\item \textbf{Action space} $A$:
Velocity $V_{control} \in  \overline{V}$, where $\overline{V} $ is the set of admissible controls,
$$V_{control} = f(X_{cam}, X_{goal}, V_{current}).$$
\end{enumerate}

Angle error - the angle between the robot's orientation and the direction from the robot to the target object.
Distance error - the distance from the robot to the target object.
The text command should provide information about the destination agent need to go to.
The control objective is to move the base from the initial position to a position with a given angle and distance error while avoiding collisions with objects inside the room and walls.  
Thus, at each moment, the robot's state satisfies $ \overline{S}$, where $ \overline{S} \subset S$ is the set of admissible (collision-free) base states.

\textbf{Robot Initialization During Training}  
Let $R \in [0,3]$ meters be the initial distance error,  
$\phi \in [0,\pi]$ be the absolute angular error,  
$\alpha \in [0, 2\pi)$ be the angle defining the robot’s position on a circle of radius $R$.  
The initialization function outputs the robot’s position $P(R,\phi, \alpha)$, where only the angle $\alpha$ is random.  
$RP$ represents the set of admissible (i.e., non-colliding with obstacles) positions of the agent in the scene.  
If the agent's position $p \notin RP$, the initialization function is restarted.  

\textbf{Curriculum Learning} 
The difficulty level consists of parameters distance error $R$ and angle error $\phi$.  
Let episode $t$ correspond to difficulty level $i$, i.e., $R(i)$ and $\phi(i)$.  
If the success rate over the last 30 episodes (excluding IL episodes) exceeds 85\% and $\phi(i)$ is not equal to the maximum deviation angle, then $\phi(i)$ is increased by a predefined value.  
Otherwise, $\phi(i)$ is set to zero, and $R(i)$ is increased.\\  
The robot initially appears close to the target area and quickly achieves a 100\% success rate.  
Then, the task is gradually made more difficult.  
A sparse reward function of the following form is used:  

\textbf{Reward Design}  
\begin{enumerate}  
    \item If none of the subtasks are completed: $reward = \frac{-2}{10}$  
    \item If only one subtask is completed: $reward = \frac{-1}{10}$  
    \item If the second subtask is completed: $reward = 2$  
    \item If the time limit is exceeded or a collision occurs: $reward = -5$  
\end{enumerate}  

\begin{figure}  
    \centering  
    \includegraphics[width=0.6\linewidth]{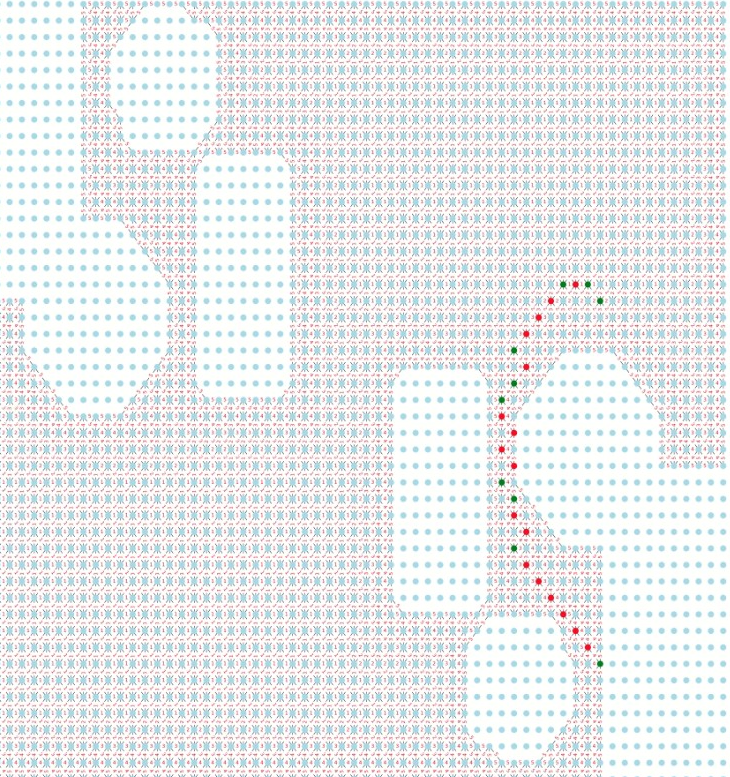}  
    \caption{An example of path construction on a grid of admissible positions using Dijkstra’s algorithm. Green points represent the simplified trajectory.}  
    \label{fig:enter-label}  
\end{figure}

\textbf{Imitation Learning}  
Paths from any grid point on the map to target positions are constructed using Dijkstra’s algorithm.  
The algorithm computes and stores paths from each grid point, constructed over $RP$, to each object position from a limited set of possible initial object positions at the beginning of training.  
The paths are simplified by removing points that lie between adjacent points arranged in a straight line or diagonal in Fig. \ref{fig:enter-label}.  
Then, at the beginning of an episode, during robot initialization, the nearest grid point to the agent is selected, and the corresponding path for the current target position is chosen.  
The agent follows this path using the Pure Pursuit algorithm \cite{coulter1992implemeyang2018visualntation}.  
The Pure Pursuit algorithm uses a target point ahead of the robot at a specified lookahead distance and calculates the trajectory curvature to follow it, minimizing the angle between the movement direction and the target point.  
The turning radius is determined based on motion geometry, allowing smooth course corrections toward the target trajectory.

\textbf{General work of the pipeline} There are two modes of action that can be taken by an agent during an episode: action, obtained from the policy, and action, obtained from their control module. The monitoring module (which only works in training mode), which implements IL, is needed in order to avoid local minima. As input, he gets the speed v of the job and the position p of the job, the goal and the obstacles.To maintain a balance between finding new strategies and avoiding local minima, it was decided to use the control module in 30\% (Mode in Fig.\ref{fig:block}) of episodes. I. e. 30\% of the episodes, the robot moves according to the control module. The SAC receives training data from the replay buffer, then the actions are recorded into reply buffer along with the speed that was obtained in the simulation and the embeddings of the current image from the cameras, the text command and the scene graph. The Fig.\ref{fig:block} shows the general block diagram of the proposed navigation algorithm leveraging 3D scene graph.







\section{Experimental results}
\subsection{Task description}
\begin{figure}[!t] 
    \centering
    \includegraphics[width=1\linewidth]{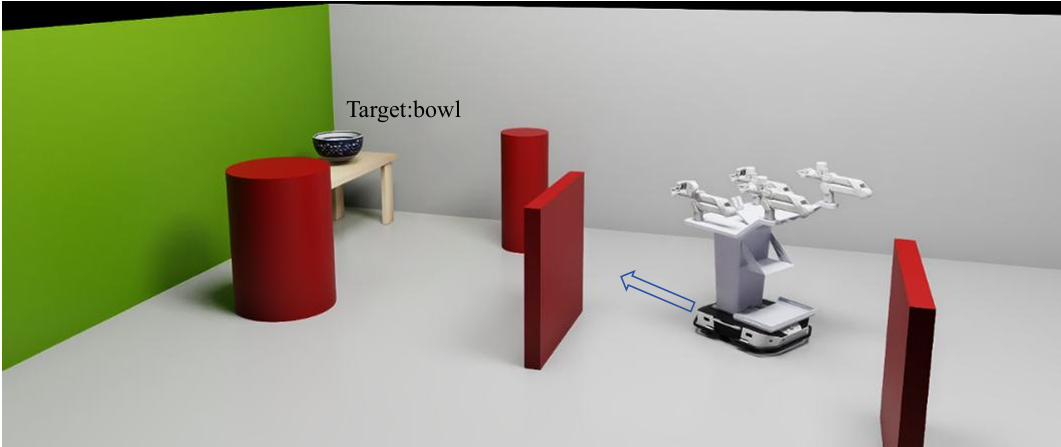}
    \caption{Aloha robot moves to the target - bowl on the table.}
    \label{fig:aloha}
\end{figure}

\textbf{Description of the Software Platform}  
The simulation is performed in the Isaac Sim environment.  
The training algorithm used is SAC (libraries: stable baselines3, gymnasium).  
The agent is the mobile wheeled robot Aloha \cite{fu2024mobile} with two RGB cameras mounted on each of its front manipulators (Fig. \ref{fig:aloha}). 
The target object is a bowl.  
The observation vector is obtained by concatenating the angular and linear velocities with CLIP embeddings of the target object and current images.  

\begin{figure}  
    \centering  
    \includegraphics[width=0.2\textwidth]{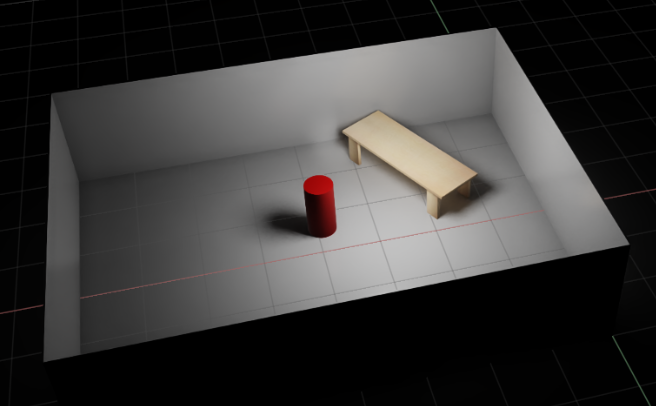} 
    \includegraphics[width=0.2\textwidth]{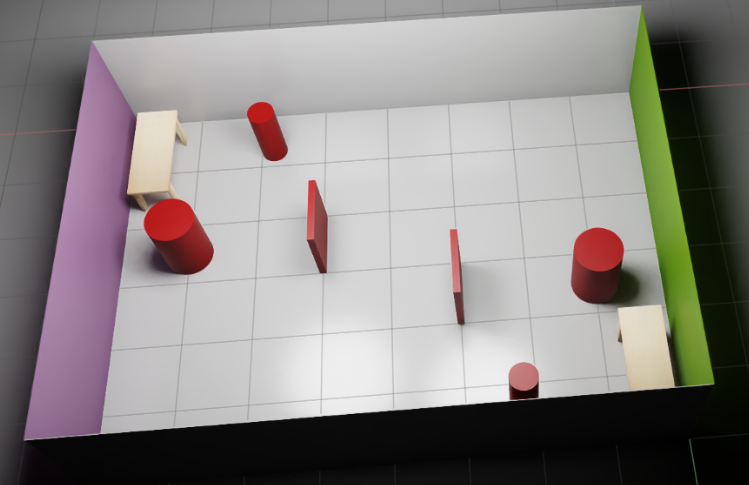}
    \caption{Experiment Scene 1 (simple) and Scene 2 (complex)}  
    \label{fig:scene1}  
\end{figure}

\textbf{How to measure success rate} 
An episode is considered successful if the angular and distance errors do not exceed 13 degrees and 1.1 meters, respectively.

\subsection{Experiment on 3D scene graph construction}  
\begin{figure}[!t]   
    \centering  
    \includegraphics[width=1\linewidth]{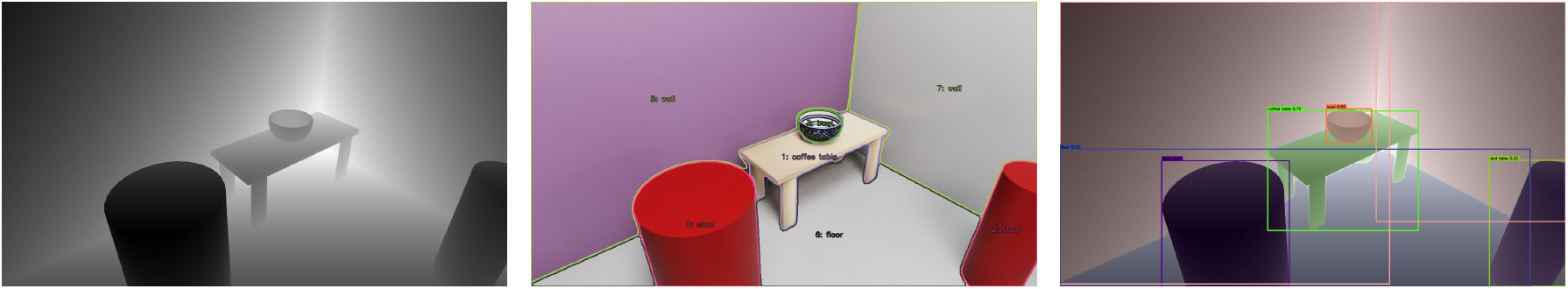}  
    \caption{ConceptGraph-YOLO \cite{gu2024conceptgraphs} visualization of a frame of camera.}  
    \label{fig:cg_yolo}  
\end{figure}

We choose ConceptGraph \cite{gu2024conceptgraphs} and BBQ \cite{linok2024barequeriesopenvocabularyobject} as the graph construction methods. The table.\ref{table:cg_bbq} shows the semantic segmentation results of the navigation scene in Fig.\ref{fig:scene1} with the target position $(7.5, 1, 0.7)m$. Fig.\ref{fig:cg_yolo} shows a ConceptGraph visualization of one frame of collected data.

\subsection{Experiment on a simple stage without the control module}  
\begin{figure}[!t]  
    \centering  
    \includegraphics[width=1\linewidth]{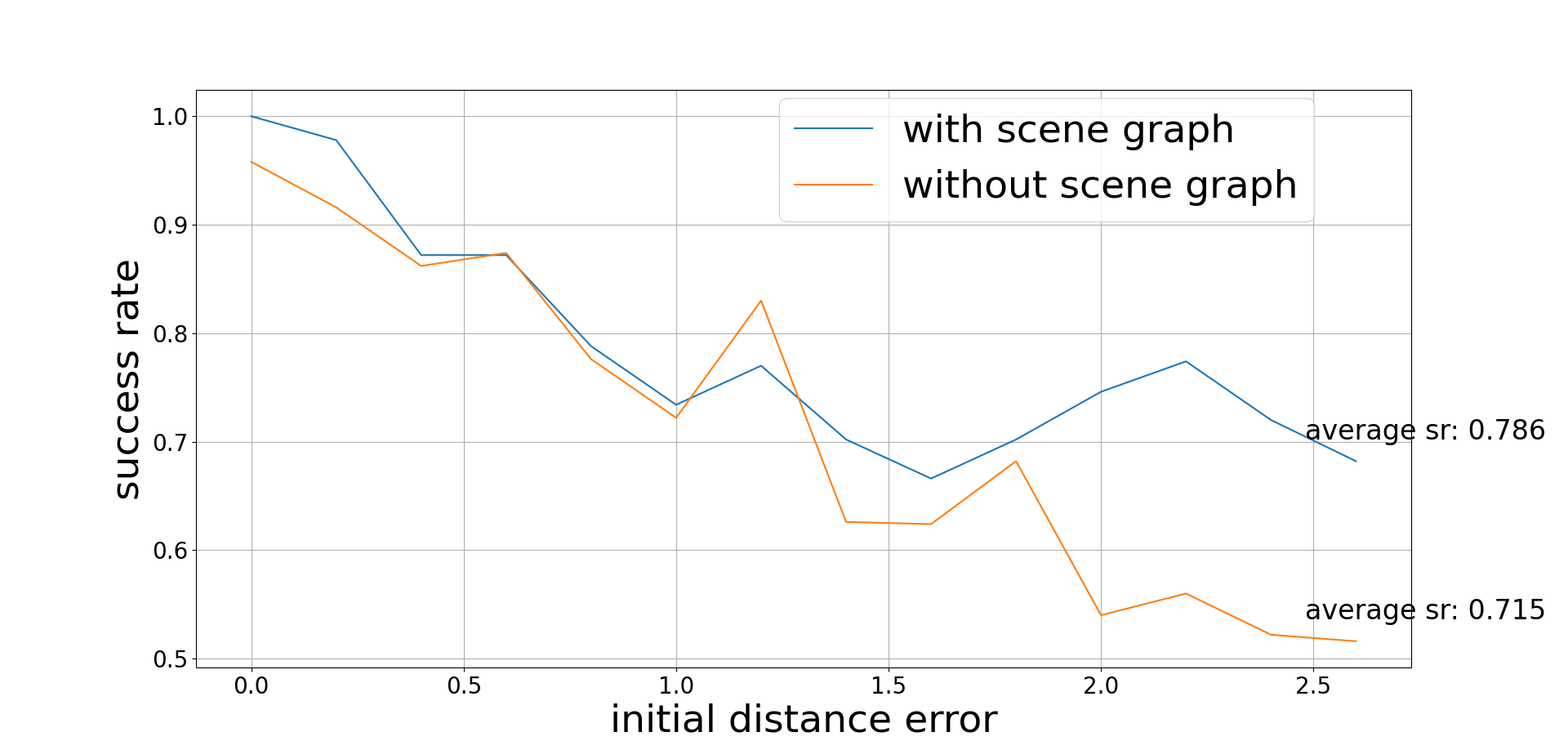}  
    \caption{Graph of success rate as a function of initial distance error during inference (Experiment on a simple stage without the control module)}  
    \label{fig:g1_2}  
\end{figure}

An experiment was conducted in a simple scene \ref{fig:scene1}, where the target object was placed on a single table in one of five positions, with an obstacle in front of the table in the form of a red pole.  
In this experiment, the control module was absent.  
The training lasted for 650,000 timesteps.  
The average success rate during inference was 78.6\% for the approach using a scene graph and 71.5\% for the approach without it as shown in Fig. \ref{fig:g1_2}.  

\begin{figure}[!t]   
    \centering  
    \includegraphics[width=1\linewidth]{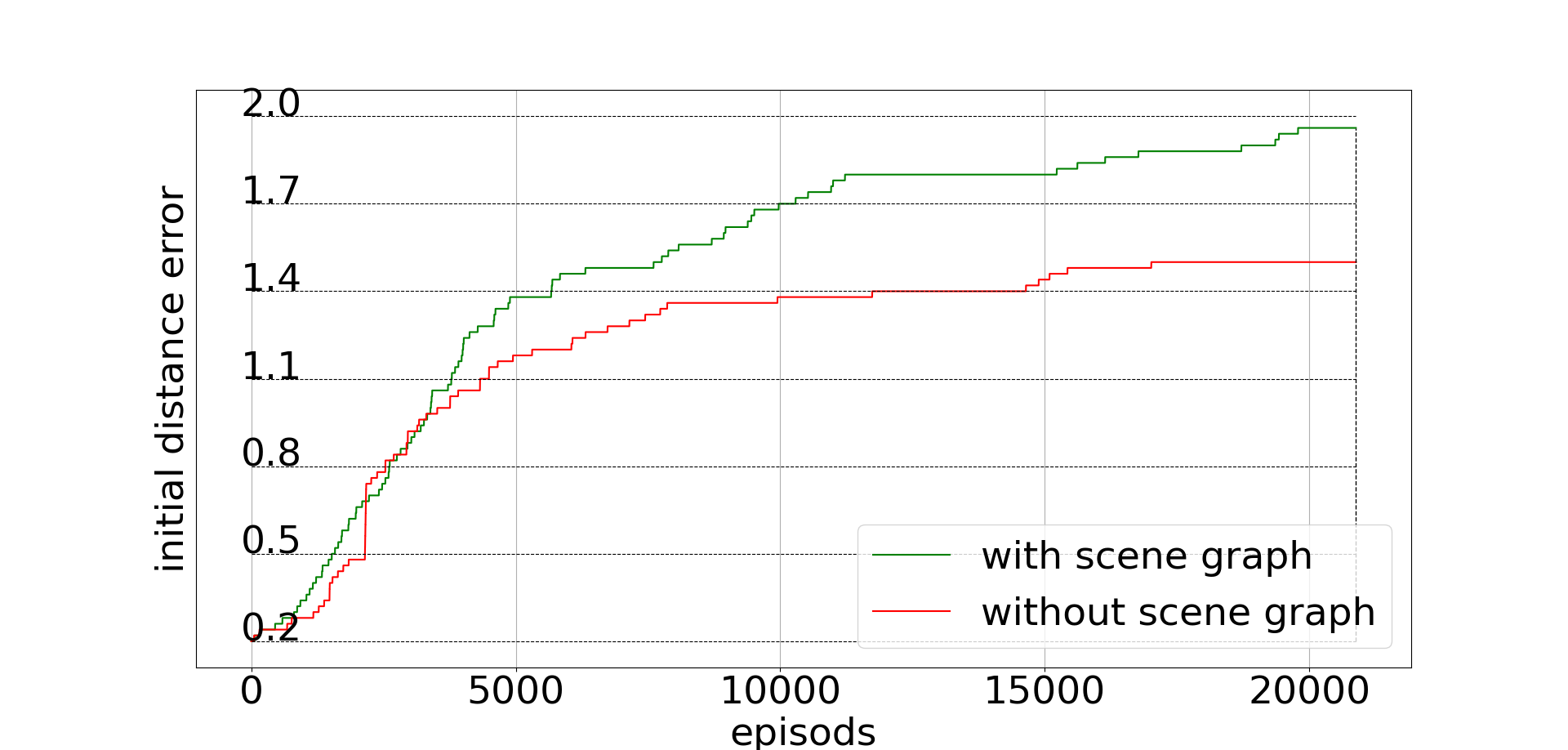}  
    \caption{Graph of the achieved difficulty level as a function of the number of episodes during training (Experiment on a simple stage without the control module)}  
    \label{fig:g1_1}  
\end{figure}  

\subsection{Experiment on a complex stage with the control module}
Next, an experiment was conducted in a more complex scene in Fig. \ref{fig:scene1} with a control module. We have two colored walls with adjacent tables. The text command includes the color of the wall where the table with the target object is located. The target object was placed on one of two tables in one of three possible positions.

In the experiment, we compared three approaches that differ in scene graph representations. In the first RL approach, the scene graph is not used. In the second SGN-CIRL (target), we use the information of target - bowl from the 3D scene graph, which includes position and extent of its 3d bounding box and its text clip embedding. In the third SGN-CIRL (scene), we use the cosine similarity value to calculate the similarity between the each object in scene and the target object - bowl. This similarity is used for weighting in the pooling operation, compressing the original (num\_object, 518) representation to 518 dimensions.

\begin{figure}[!t]  
    \centering  
    \includegraphics[width=1\linewidth]{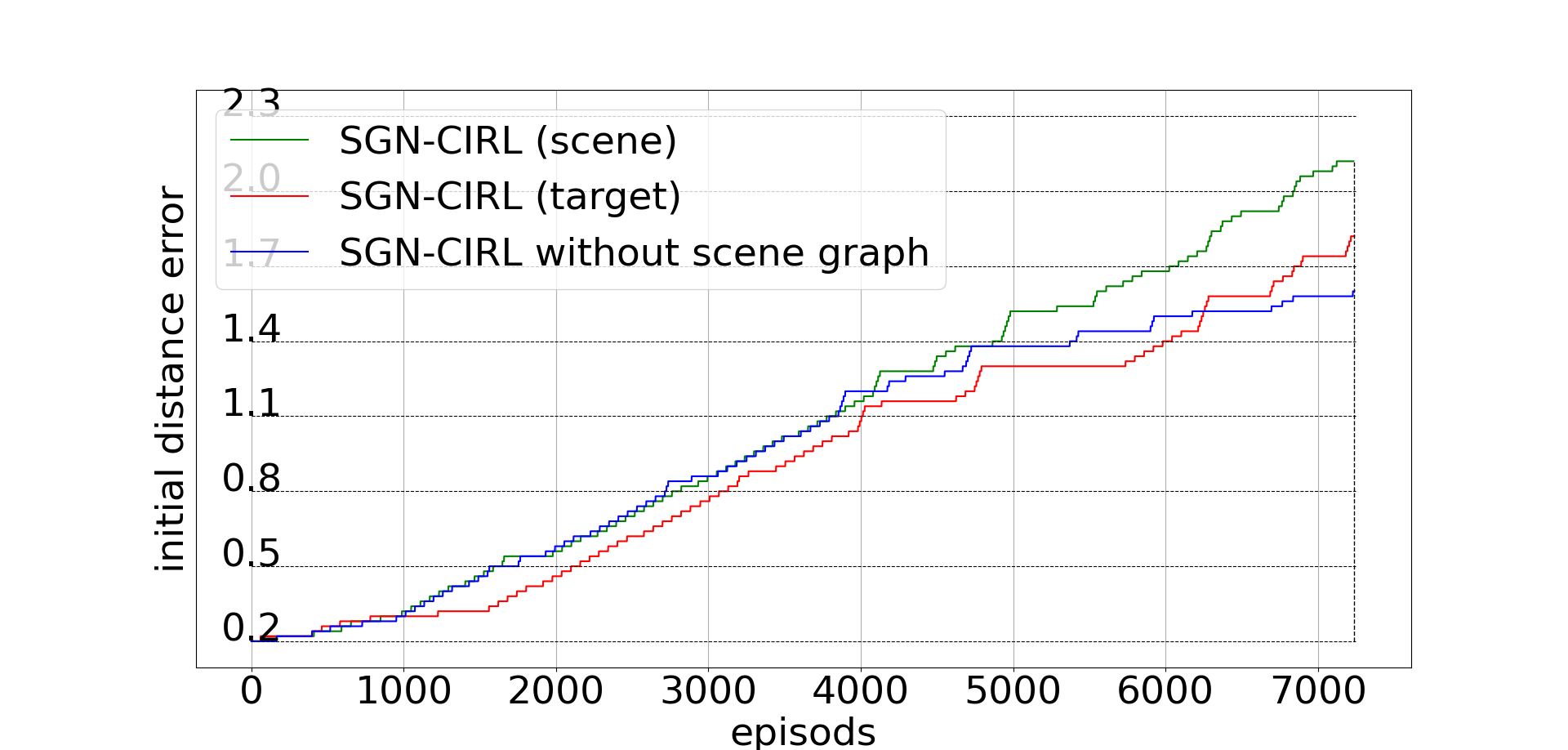}  
    \caption{Graph of the achieved difficulty level as a function of the number of episodes during training (Experiment on a complex stage with the control module)}  
    \label{fig:g2_1}  
\end{figure}

\begin{figure}[!t]  
    \centering  
    \includegraphics[width=1\linewidth]{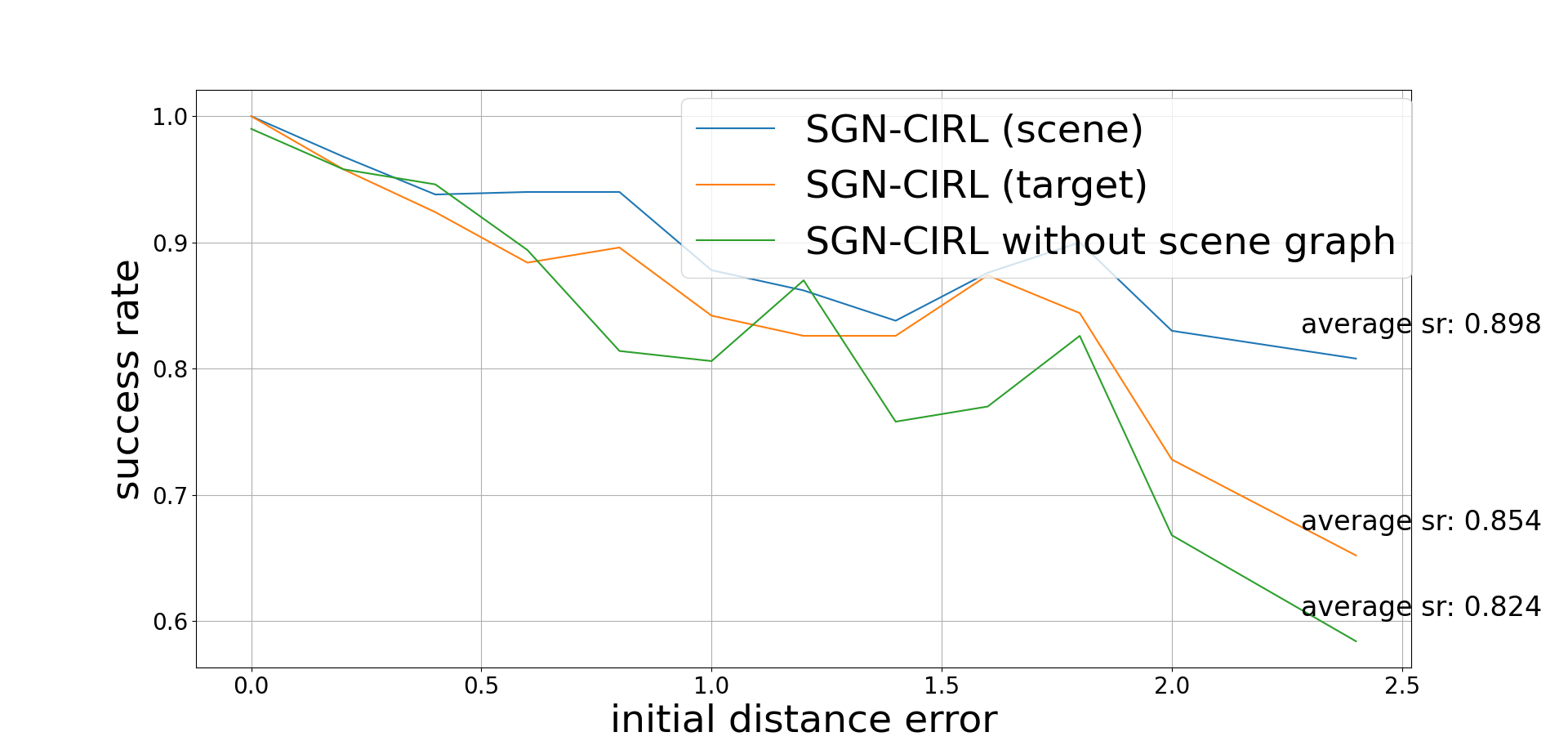}  
    \caption{Graph of success rate as a function of initial distance error during inference (Experiment on a complex stage with the control module)}  
    \label{fig:g2_2}  
\end{figure}

The training lasted for 500,000 timesteps.
The difference in learning rate is shown in Fig.~\ref{fig:g2_1}
The average success rate during inference was 85.4\% for the SGN-CIRL (target) approach, 89.8\% for the SGN-CIRL (scene) approach and 82.4\% for the RL approach without, what is shown in Fig.~\ref{fig:g2_2}. 
\begin{table}[t]
\centering
\caption{SEMANTIC SEGMENTATION ON NAVIGATION SCENE.}
\label{table:cg_bbq}
\begin{tabular}{lcccc}
\hline
\textbf{Methods} & \textbf{mIoU$\uparrow$} & \textbf{mRecall$\uparrow$} & \textbf{mPrecision$\uparrow$} & \textbf{fmiou$\uparrow$} \\
\hline
BBQ-CLIP & 0.58 & 0.70 & 0.60 & 0.58 \\
CG-YOLO & \textbf{0.96} & \textbf{0.97} & \textbf{0.98} & \textbf{0.99} \\
\hline
\end{tabular}
\end{table}

\begin{table}[!t]
    \centering
    \caption{Comparison of experimental results}
    \label{tab:results}
    \begin{tabular}{p{3.0cm}|p{2.2cm}|p{2.2cm}}
    \toprule
    \textbf{Model} & \textbf{Success rate (Init. dist. error 2 m)} & \textbf{Success rate (Init. dist. error 2.4 m)} \\
    \midrule
    \multicolumn{3}{l}{\textit{Experiment on a simple stage without the control module}}\\
    \hline
    RL with scene graph    & \textbf{75\%} (+20\%)  & \textbf{71\%} (+19\%)\\
    RL without scene graph & 55\%  & 52\% \\
    \midrule
    \multicolumn{3}{l}{\textit{Experiment on a complex stage with the control module}}\\
    \hline
    SGN-CIRL (scene)     & \textbf{84\% (+17\%)}  & \textbf{81\% (+22\%)} \\
    SGN-CIRL (target)    &      72\% (+5\%)  & 66\% (+7\%)\\
    SGN-CIRL without scene graph & 67\%  & 59\% \\
    \bottomrule
    \end{tabular}
    \label{table}
\end{table}
\subsection{Experiment with randomly placed obstacles}
\begin{figure}
    \centering
    \includegraphics[width=0.8\linewidth]{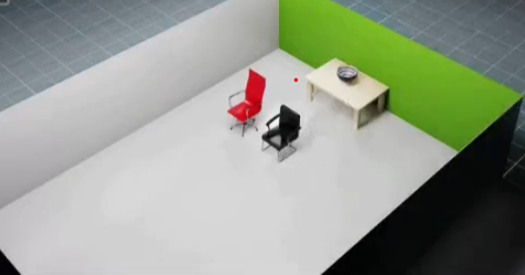}
    \caption{environment with randomly placed obstacles}
    \label{fig:2chair}
\end{figure}

In this experiment, we complicated the appearance of the obstacle by swapping abstract shapes for chairs in two colors (red and black) and a random arrangement of 9 different configurations (Fig.~\ref{fig:2chair}), from 0 to 3 chairs on stage. Just for the sake of experimental purity, the scene graphs were generated manually (i.e. without using the bbq algorithm, SGN-CIRL (ground truth graphs)). Thus, we have the following results in the table 3, which show the inability to solve the problem without the knowledge graph and sufficient success with its use.

\begin{table}[!t]
    \centering
    \caption{Comparison of randomly placed obstacles experimental results}
    \label{tab:results2}
    \begin{tabular}{p{5.0cm}|p{2.2cm}}
    \toprule
    \textbf{Model} & \textbf{Success rate (Init. dist. error 3 m)} \\
    \hline
    RL with scene graph    & \textbf{21\%} (+20\%)\\
    \hline
    SGN-CIRL (ground truth graphs)     & \textbf{82\% (+61\%)}\\
    \bottomrule
    \end{tabular}
    \label{table}
\end{table}

\section{Discussion}
We can see that in both cases, although SGN-CIRL approach with the 3D Scene Graph  slightly improves the results of the graph-free approach on average, in difficult to navigate cases the gain of the first approach starts to increase and is around 10-20\%, what is shown in Table ~\ref{table}.

In future work, we will complicate the scenes, increase their number, the number of targets, and train the models for longer. It is also possible to switch to dynamic scenes.

\section{Conclusion}

This paper presents an original framework named SGN-CIRL for mapless navigation that leverages 3D scene graphs as a core component for reinforcement learning. By integrating the scene graph construction approach, the proposed method enables the agent to model spatial relationships between objects and predict target locations, enhancing navigation efficiency in partially observable environments. The experiments conducted in Isaac Sim demonstrate that incorporating scene graphs significanyly improves navigation success rate, especially in complex scenarios where traditional RL methods struggle with generalization.

A key contribution of this work is the integration of scene graphs with reinforcement learning, imitation learning, and curriculum learning to create a robust and adaptive robot navigation strategy. Experimental results show that agents utilizing scene graphs achieve higher success rates compared to baselines without such representations, particularly in scenarios involving occlusions and ambiguous spatial layouts.

In the future, we will focus on scaling the approach to dynamic environments, incorporating moving objects and additional sensory modalities. A critical direction for further research is the optimization of the scene graph representation, including the selection of the most informative embedding for objects and relationships. Fine-tuning the parameters of the scene graph, such as node connectivity, feature selection, and graph update frequency, may further enhance its effectiveness in guiding the agent’s decision-making process. Additionally, exploring alternative architectures for integrating graph information into reinforcement learning models could lead to more efficient and generalizable navigation strategies.

The results highlight the potential of a new approach, knowledge-based reinforcement learning, for efficient and adaptive autonomous navigation, while also opening avenues for deeper exploration of structured knowledge representations in robotic decision-making.



\addtolength{\textheight}{-12cm}   




\bibliographystyle{IEEEtran} 
\bibliography{ref} 

\end{document}